%% file: main.tex
\documentclass[10pt,twocolumn,letterpaper]{article}
\usepackage[pagenumbers]{iccv}              % To produce the CAMERA-READY version
\input{preamble}

\definecolor{iccvblue}{rgb}{0.21,0.49,0.74}
\usepackage[pagebackref,breaklinks,colorlinks,allcolors=iccvblue]{hyperref}

\title{Bridging the Skeleton-Text Modality Gap: Diffusion-Powered\\Modality Alignment for Zero-shot Skeleton-based Action Recognition}

\author{Jeonghyeok Do \qquad\qquad Munchurl Kim\thanks{Corresponding author.}\\
[0.7em]
Korea Advanced Institute of Science and Technology\\
{\tt\small \{ehwjdgur0913, mkimee\}@kaist.ac.kr}\\
\small{\url{https://kaist-viclab.github.io/TDSM_site}}
}

\begin{document}
\maketitle
\input{sec/0_abstract}    
\input{sec/1_intro}
\input{sec/2_related}
\input{sec/3_preliminary}
\input{sec/4_method}
\input{sec/5_experiment}

\input{sec/6_conclusion}

\vspace{0.2cm}
\noindent \textbf{Acknowledgements.}\quad This work was supported by IITP grant funded by the Korea government(MSIT) (No.RS2022-00144444, Deep Learning Based Visual Representational Learning and Rendering of Static and Dynamic Scenes).

\clearpage

\input{sec/X_suppl}

{
    \small
    \bibliographystyle{ieeenat_fullname}
    \bibliography{main}
}

\end{document}

%% file: preamble.tex
\usepackage{graphicx}
\usepackage{booktabs}
\usepackage{amsmath}
\usepackage{amssymb}
\usepackage{multirow}
\usepackage{array}
\usepackage{makecell}
\usepackage{bm}

%% file: sec/0_abstract.tex
\begin{abstract}

In zero-shot skeleton-based action recognition (ZSAR), aligning skeleton features with the text features of action labels is essential for accurately predicting unseen actions. ZSAR faces a fundamental challenge in bridging the modality gap between the two-kind features, which severely limits generalization to unseen actions. Previous methods focus on direct alignment between skeleton and text latent spaces, but the modality gaps between these spaces hinder robust generalization learning. Motivated by the success of diffusion models in multi-modal alignment (e.g., text-to-image, text-to-video), we firstly present a diffusion-based skeleton-text alignment framework for ZSAR. Our approach, Triplet Diffusion for Skeleton-Text Matching (TDSM), focuses on cross-alignment power of diffusion models rather than their generative capability. Specifically, TDSM aligns skeleton features with text prompts by incorporating text features into the reverse diffusion process, where skeleton features are denoised under text guidance, forming a unified skeleton-text latent space for robust matching. To enhance discriminative power, we introduce a triplet diffusion (TD) loss that encourages our TDSM to correct skeleton-text matches while pushing them apart for different action classes. Our TDSM significantly outperforms very recent state-of-the-art methods with significantly large margins of 2.36\%-point to 13.05\%-point, demonstrating superior accuracy and scalability in zero-shot settings through effective skeleton-text matching.

\end{abstract}

\vspace{-0.3cm}

%% file: sec/1_intro.tex
\section{Introduction}
\label{sec:intro}

Human action recognition \cite{Survey1, Survey2, Survey3, Survey4} focuses on classifying actions from movements, with RGB videos commonly used due to their accessibility. However, recent advancements in depth sensors \cite{Kinetics} and pose estimation algorithms \cite{Openpose, HRNet} have driven the adoption of skeleton-based action recognition. Skeleton data offers several advantages: it captures only human poses without background noise, ensuring a compact representation. Furthermore, 3D skeletons remain invariant to environmental factors such as lighting, background, and camera angles, providing consistent 3D coordinates across conditions \cite{PoseC3D}.

\begin{figure}[tbp]
  \centering
  \includegraphics[width=1.0\columnwidth]{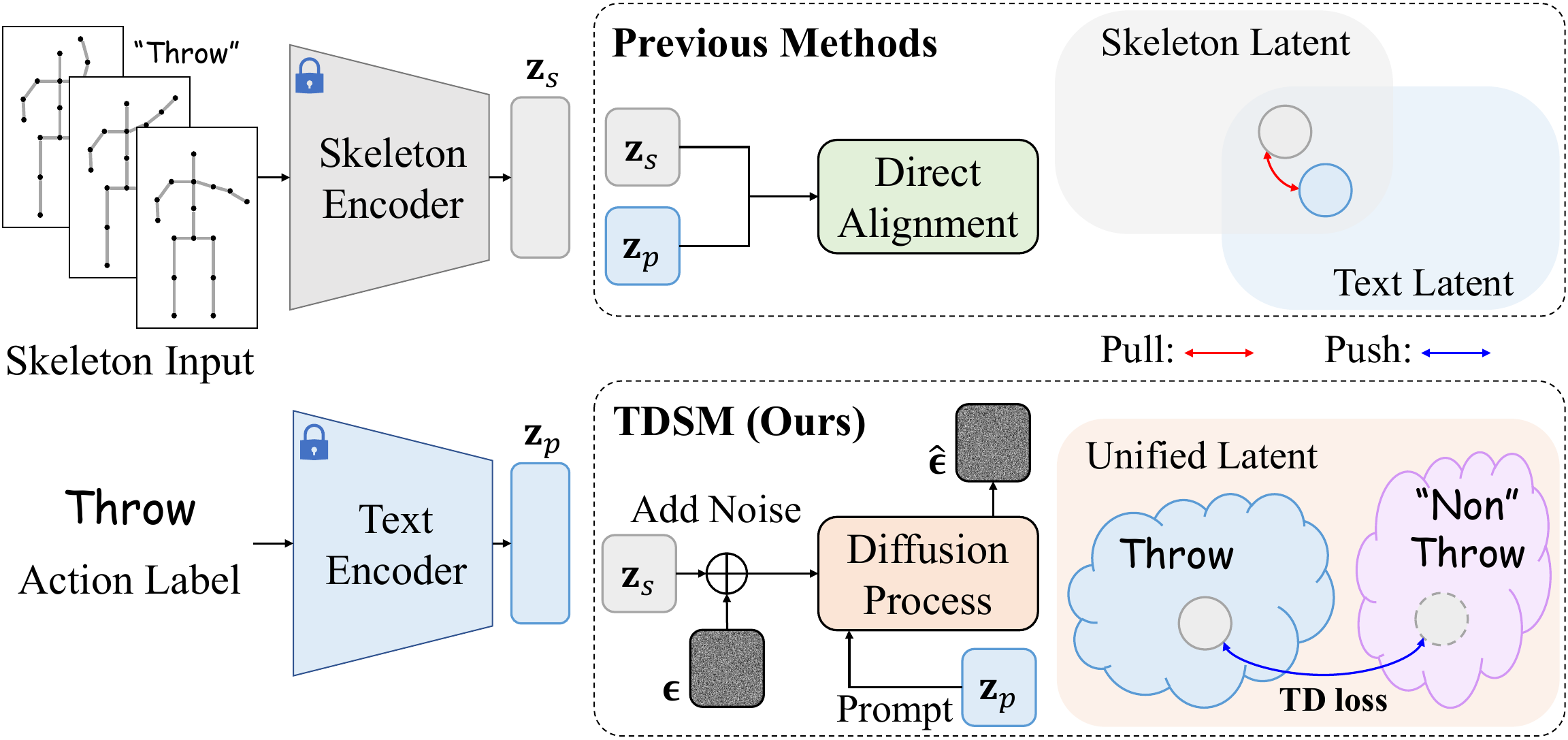}
  \caption{Overview of our Triplet Diffusion for Skeleton-Text Matching (TDSM) pipeline versus previous methods. While the previous methods rely on direct alignment between skeleton and text latent spaces, thus suffering from modality gaps that limit generalization, our TDSM pipeline overcomes this challenge by utilizing the cross-modality alignment power of diffusion models, establishing a more unified and robust skeleton-text representation for effective cross-modal matching.}
  \label{fig:motiv}
  \vspace{-0.6cm}
\end{figure}

Despite these benefits, the fully supervised skeleton-based action recognition methods \cite{STGCN, ShiftGCN, PoseC3D, BlockGCN, SkateFormer, FRHead, CTRGCN, InfoGCN} tend to perform well, but annotating every possible action is impractical for a large number of possible action classes. In addition, retraining models for new classes incurs a significant cost. So, zero-shot skeleton-based action recognition (ZSAR) \cite{CADAVAE, SynSE, SMIE, PURLS, SADVAE, STAR, InfoCPL, DVTA, MSF} addresses this issue by enabling predictions for unseen actions without requiring explicit training data, making it valuable for applications such as surveillance, robotics, and human-computer interaction, where continuous learning is infeasible \cite{ZeroSurvey1, ZeroSurvey2}. Importantly, ZSAR is possible because human actions often share common skeletal movement patterns across related actions. By leveraging these shared patterns, ZSAR methods align pre-learned skeleton features with text-based action descriptions, allowing the models to extrapolate from seen actions to unseen ones. This alignment-based approach reinforces the model’s discriminative power, ensuring scalability and reliable zero-shot recognition in real-world scenarios. However, achieving the effective alignment between skeleton data and text features entails significant challenges. While skeleton data captures temporal and spatial motion patterns, the text descriptions for action labels carry high-level semantic information. This modality gap makes it difficult to align their corresponding latent spaces effectively, thus hindering the generalization learning for unseen actions.

Diffusion models \cite{SD, SD3} have demonstrated strong cross-modal alignment capabilities by incorporating conditioning signals such as text, images, audio, or video to guide the generative process. This conditioning mechanism enables precise \textit{cross-modality alignment}, ensuring that generated outputs adhere closely to the given condition. Inspired by this property, we propose a novel framework: Triplet Diffusion for Skeleton-Prompt Matching (TDSM), which \textit{firstly} adopts diffusion models to ZSAR by conditioning the denoising process on text prompts. Our approach utilizes the reverse diffusion process to implicitly align skeleton and text features within a shared latent space, overcoming the challenges of direct feature space alignment.

Fig.~\ref{fig:motiv} illustrates the key differences between previous methods and our proposed method. The previous methods \cite{CADAVAE, SynSE, SMIE, PURLS, SADVAE, STAR, InfoCPL, DVTA, MSF} attempt to directly align skeleton and text features within separate latent spaces. However, this approach struggles with generalization due to the inherent modality gap between skeletal motion and textual semantics. On the other hand, our TDSM leverages a reverse diffusion training scheme to \textit{implicitly} align skeleton features with their corresponding text prompts, producing discriminatively fused representations within a unified latent space. More specifically, our TDSM learns to denoise noisy skeleton features conditioned on the corresponding text prompts, embedding the prompts into a unified skeleton-text latent space to better capture the semantic meaning of action labels. This implicit alignment mitigates the limitations of direct latent space mapping while enhancing robustness. Additionally, we introduce a triplet diffusion (TD) loss, which encourages tighter alignment for correct skeleton-text pairs and pushes apart incorrect ones, further improving the model’s discriminative power. As an additional benefit, the stochastic nature of diffusion process, driven by the random noise added during training, acts as a natural regularization mechanism. This prevents overfitting and enhances the model’s ability to generalize effectively to unseen actions. Our contributions are threefold:

\begin{itemize}
\item We \textit{firstly} present a diffusion-based action recognition with zero-shot learning for skeleton inputs, called a Triplet Diffusion for Skeleton-Text Matching (TDSM) which is the first framework to apply diffusion models and to implicitly align the skeleton features with text prompts (action labels) by fully taking the advantage of excellent text-image correspondence learning in generative diffusion process, thus being able to learn fused discriminative features in a unified latent space.
\item We introduce a reformulated triplet diffusion (TD) loss to enhance the model's discriminative power by ensuring accurate denoising for correct skeleton-text pairs while suppressing it for incorrect pairs.
\item Our TDSM \textit{significantly} outperforms the very recent state-of-the-art (SOTA) methods with large margins of 2.36\%-point to 13.05\%-point across multiple benchmarks, demonstrating scalability and robustness under various seen-unseen split settings.
\end{itemize}

%% file: sec/2_related.tex
\section{Related Work}
\label{sec:related}

\subsection{Zero-shot Skeleton-based Action Recognition}
\label{sec:22}

Zero-shot Skeleton-based Action Recognition (ZSAR) aims to recognize human actions from skeleton sequences without requiring labeled training data for unseen action categories. Most of the existing works focus on aligning the skeleton latent space with the text latent space. These approaches can be categorized broadly into VAE-based methods \cite{CADAVAE, SynSE, MSF, SADVAE} and contrastive learning-based methods \cite{SMIE, PURLS, STAR, DVTA, InfoCPL}.\\
\textbf{VAE-based.} The previous work, CADA-VAE \cite{CADAVAE}, leverages VAEs \cite{VAE} to align skeleton and text latent spaces, ensuring that each modality’s decoder can generate useful outputs from the other’s latent representation. SynSE \cite{SynSE} refines this by introducing separate VAEs for verbs and nouns, improving the structure of the text latent space. MSF \cite{MSF} extends this approach by incorporating action and motion-level descriptions to enhance alignment. SA-DVAE \cite{SADVAE} disentangles skeleton features into semantic-relevant and irrelevant components, aligning text features exclusively with relevant skeleton features for improved performance.\\
\textbf{Contrastive learning-based.} Contrastive learning-based methods align skeleton and text features through positive and negative pairs \cite{SimCLR}. SMIE \cite{SMIE} concatenates skeleton and text features, and applies contrastive learning by treating masked skeleton features as positive samples and other actions as negatives. PURLS \cite{PURLS} incorporates GPT-3 \cite{GPT} to generate text descriptions based on body parts and motion evolution, using cross-attention to align text descriptions with skeleton features. STAR \cite{STAR} extends this idea with GPT-3.5 \cite{GPT}, generating text descriptions for six distinct skeleton groups, and introduces learnable prompts to enhance alignment. DVTA \cite{DVTA} introduces a dual alignment strategy, performing direct alignment between skeleton and text features, while also generating augmented text features via cross-attention for improved alignment. InfoCPL \cite{InfoCPL} strengthens contrastive learning by generating 100 unique sentences per action label, enriching the alignment space.

While most existing methods rely on direct alignment between skeleton and text latent spaces, they often struggle with generalization due to inherent differences between the two modalities. In contrast, our TDSM leverages diffusion models for \textit{alignment} rather than generation. By conditioning the reverse diffusion process on action labels, we guide the denoising of skeleton features to implicitly align them with their corresponding semantic contexts of action labels. This enables more robust skeleton-text matching and improves generalization to unseen actions.

% While most existing methods rely on direct alignment between skeleton and text latent spaces, they often struggle with generalization due to inherent differences between the two modalities. To address these limitations, our TDSM leverages a novel triplet diffusion approach. It uses text prompts to guide the reverse diffusion process, embedding these prompts into the unified skeleton-text latent space for more effective implicit alignment via feature fusion. Also, our triplet diffusion loss, inspired by the triplet loss \cite{Triplet}, further enhances the model’s discriminative power by promoting correct alignments and penalizing incorrect ones.

\begin{figure*}[tbp]
  \centering
  \includegraphics[width=1.0\textwidth]{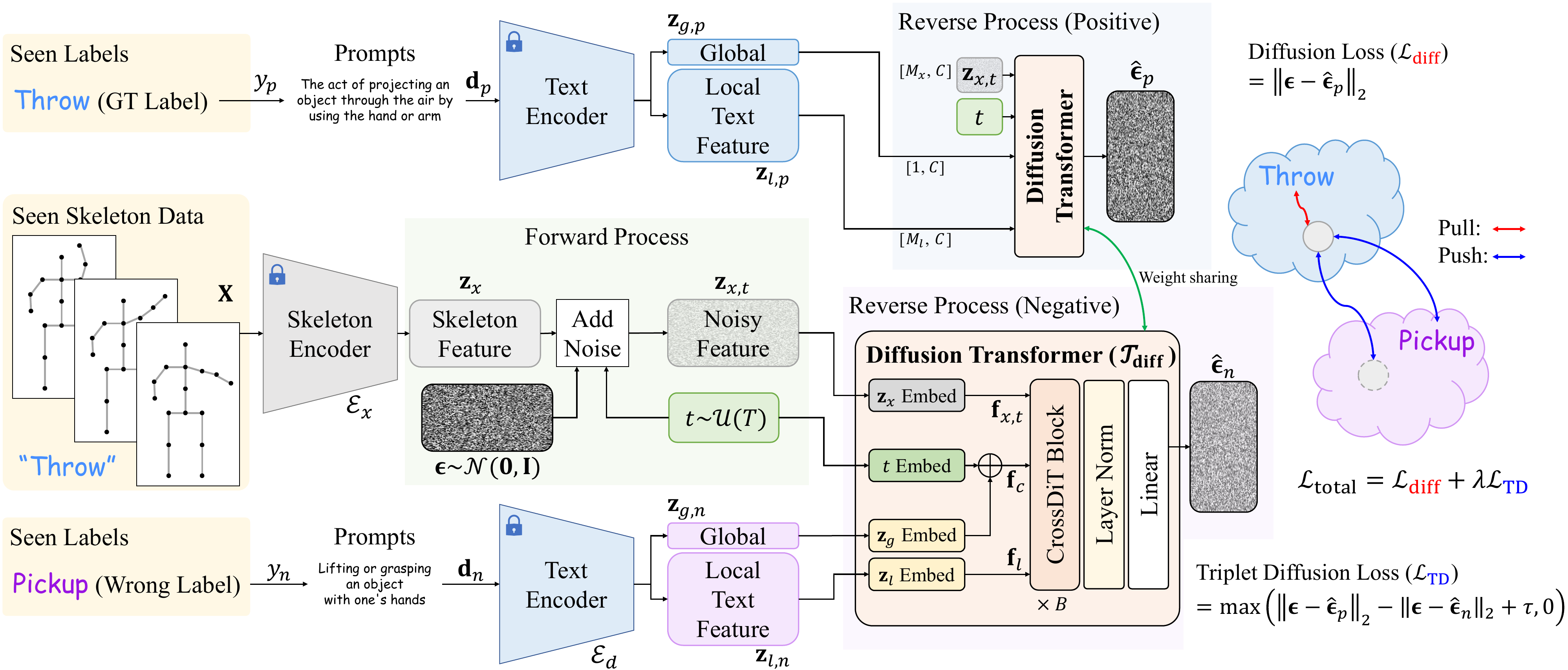}
  \vspace{-0.6cm}
  \caption{Training framework of our TDSM for zero-shot skeleton-based action recognition.}
  \label{fig:train}
  \vspace{-0.3cm}
\end{figure*}

\subsection{Diffusion Models}
\label{sec:23}

Diffusion models have become a fundamental to generative tasks by learning to reverse a noise-adding process for original data recovery. Denoising Diffusion Probabilistic Models (DDPMs) \cite{DDPM} introduced a step-by-step denoising framework, enabling the modeling of complex data distributions and establishing the foundation for diffusion-based generative models. Building on this, Latent Diffusion Models (LDMs) \cite{SD, SD3} improve computational efficiency by operating in lower-dimensional latent spaces while maintaining high-quality outputs. LDMs have been successful in various generation tasks (e.g., text-to-image, text-to-video, image-to-video), showing the potential of diffusion models for cross-modal alignment tasks. In the reverse diffusion process, LDMs employ a denoising U-Net \cite{Unet} where text prompts are integrated with image features through cross-attention blocks, effectively guiding the model to align the two modalities. Further extending this line of research, Diffusion Transformers (DiTs) \cite{DiT} integrate transformer architectures into diffusion processes. In this work, we leverage the aligned fusion capabilities of diffusion models, focusing on the learning process during reverse diffusion rather than their generative power. Specifically, we utilize a DiT-based network as a denoising model, where text prompts guide the denoising of noisy skeleton features. This approach embeds text prompts into the unified latent space in the reverse diffusion process, ensuring robust fusion of the two modalities and enabling effective generalization to unseen actions in zero-shot recognition settings.

\noindent\textbf{Zero-shot tasks with diffusion models.} Recently, diffusion models have also been extended to zero-shot tasks in RGB-based vision applications, such as semantic correspondence \cite{zhang2024tale}, segmentation \cite{tian2024diffuse, burgert2022peekaboo}, image captioning \cite{yang2023ig}, and image classification \cite{clark2024text, li2023your}. These approaches often rely on large-scale pretrained diffusion models, such as LDMs \cite{SD, SD3}, trained on datasets like LAION-5B \cite{LAION} with billions of text-image pairs. In contrast, our approach demonstrates that diffusion models can be effectively applied to smaller, domain-specific tasks, such as skeleton-based action recognition, without the need for large-scale finetuning. This highlights the versatility of diffusion models beyond large-scale vision-language tasks, providing a practical solution to zero-shot generalization with limited data resources.

%% file: sec/3_preliminary.tex
\section{Preliminaries: Diffusion Process}
\label{sec:pre}

Diffusion models \cite{DDPM, SD, DiT} are a class of generative models that progressively denoise a noisy sample to generate a target distribution. The diffusion process \cite{DDPM} consists of two main stages: a forward process and a reverse process.

\noindent \textbf{Forward diffusion process.} In the forward process, Gaussian noise is incrementally added to the target data $\mathbf{x}_0$ over discrete timesteps $t$, gradually transforming it into a Gaussian noise distribution $\bm{\epsilon} \sim \mathcal{N}(\mathbf{0}, \mathbf{I})$. This can be formulated as: $q(\mathbf{x}_t | \mathbf{x}_{t-1}) = \mathcal{N}(\mathbf{x}_t; \sqrt{1 - \beta_t} \mathbf{x}_{t-1}, \beta_t\mathbf{I})$,  where $q(\mathbf{x}_t | \mathbf{x}_{t-1})$ follows a Markov chain that progressively corrupts $\mathbf{x}_0$ into noise, and $\beta_t$ controls the noise schedule. By reparameterizing the forward process, $\mathbf{x}_t$ can be directly expressed in terms of $\mathbf{x}_0$ as: $\mathbf{x}_t = \sqrt{\bar{\alpha}_t} \mathbf{x}_0 + \sqrt{1 - \bar{\alpha}_t} \bm{\epsilon}$, where $\bar{\alpha}_{t} = \prod_{s=1}^{t} (1 - \beta_{s})$ controls the noise level at step $t$.

\noindent \textbf{Reverse diffusion process.} The reverse process learns to recover the original data $\mathbf{x}_0$ from a noisy sample by estimating the denoising step conditioned on previous timesteps as: $p_\theta(\mathbf{x}_{t-1} | \mathbf{x}_t) = \mathcal{N}(\mathbf{x}_{t-1}; \bm{\mu}_\theta(\mathbf{x}_t, t), \mathbf{\Sigma}_\theta(\mathbf{x}_t, t))$, where $p_\theta(\mathbf{x}_{t-1} | \mathbf{x}_t)$ follows a Gaussian conditional distribution that gradually reconstructs $\mathbf{x}_0$ by estimating the denoised mean $\bm{\mu}_\theta(\mathbf{x}_t, t)$ and covariance $\mathbf{\Sigma}_\theta(\mathbf{x}_t, t)$. Here, $\bm{\mu}_\theta$ and $\mathbf{\Sigma}_\theta$ are predicted by a neural network trained to approximate the inverse of the forward noise addition, enabling the model to progressively refine $\mathbf{x}_t$ into a clean representation.

%$ where $\bm{\mu}_\theta$ and $\mathbf{\Sigma}_\theta$ are predicted by a neural network trained to approximate the inverse of the forward noise addition.

\noindent \textbf{Objective function.} To train the reverse process, the objective function is derived by minimizing the variational bound on the negative log-likelihood $\mathbb{E}\left[-\log p_\theta(\mathbf{x}_{0}) \right]$. It can be simplified by reparameterizing $\bm{\mu}_\theta$ as a noise prediction network $\epsilon_\theta$, leading to the following loss as: $\mathcal{L}_{\text{diff}}=\| \epsilon_\theta (\mathbf{x}_t, t) - \bm{\epsilon} \|_2$.

\noindent \textbf{Cross-modality conditioning in diffusion models.} An important property of diffusion models is their ability to incorporate conditioning signals $\mathbf{c}$ such as text, images, audio, or video to guide the generative process. This conditioning mechanism enables strong \textit{cross-modality alignment}, allowing the models to generate outputs that closely align with the given conditions $\mathbf{c}$. To integrate conditioning into the diffusion process, the noise prediction model $\epsilon_\theta$ is conditioned on $\mathbf{c}$, modifying the objective function as follows: $\mathcal{L}_{\text{diff}}=\| \epsilon_\theta (\mathbf{x}_t, t; \mathbf{c}) - \bm{\epsilon} \|_2$. This formulation ensures that the denoising process is guided by the conditions $\mathbf{c}$, enforcing the alignment between the generated output and the conditioning signals $\mathbf{c}$.

%% file: sec/4_method.tex
\section{Methods}
\label{sec:method}

\subsection{Overview of TDSM}
In the training phase, we are given a dataset 
\begin{equation}
    \mathcal{D}_{\text{train}} = \{(\mathbf{X}_{i}, y_{i})\}_{i=1}^{N},   y_{i} \in \mathcal{Y},
\end{equation}
where $\mathbf{X}_{i} \in \mathbb{R}^{T \times V \times M \times C_{\text{in}}}$ represents a skeleton sequence, and $y_{i}$ is the corresponding ground truth label. Each skeleton sequence $\mathbf{X}_{i}$ consists of sequence length $T$, the number $V$ of joints, the number $M$ of actors, and the dimensionality $C_{\text{in}}$ representing each joint. The label $y_{i}$ belongs to the set of seen class labels $\mathcal{Y}$. Here, $N$ denotes the total number of training samples in the seen dataset. In the inference phase, we are provided with a test dataset 

\begin{equation}
    \mathcal{D}_{\text{test}} = \{(\mathbf{X}_{j}^{u}, y_{j}^{u})\}_{j=1}^{N_{u}},   y_{j}^{u} \in \mathcal{Y}_{u},
\end{equation}

\noindent where $\mathbf{X}_{j}^{u} \in \mathbb{R}^{T \times V \times M \times C_{\text{in}}}$ denotes skeleton sequences from unseen classes, and $y_{j}^{u}$ are their corresponding labels. In this phase, $N_{u}$ represents the total number of test samples from unseen classes. In the zero-shot setting, the seen and unseen label sets are disjoint, i.e., 

\begin{equation}
    \mathcal{Y} \cap \mathcal{Y}_u = \varnothing.
\end{equation}

\noindent We train the TDSM using $\mathcal{D}_{\text{train}}$ and enable it to generalize to unseen classes from $\mathcal{D}_{\text{test}}$. By learning a robust discriminative fusion of skeleton features and text descriptions, the model can predict the correct label ${\hat{y}}^{u} \in \mathcal{Y}_{u}$ for an unseen skeleton sequence $\mathbf{X}_{j}^{u}$ during inference.

Fig.~\ref{fig:train} provides an overview of our training framework of TDSM. As detailed in Sec.~\ref{sec:32}, the pretrained skeleton encoder $\mathcal{E}_{x}$ and text encoder $\mathcal{E}_{d}$ embed the skeleton inputs $\mathbf{X}$ and prompt input $\mathbf{d}$ with an action label $y$ into their respective feature spaces, producing the skeleton feature $\mathbf{z}_{x}$ and two types of text features: the global text feature $\mathbf{z}_{g}$ and the local text feature $\mathbf{z}_{l}$. The skeleton feature $\mathbf{z}_{x}$ undergoes the forward process, where noise $\bm{\epsilon}$ is added to it. In the reverse process, as described in Sec.~\ref{sec:33}, the Diffusion Transformer $\mathcal{T}_\text{diff}$ serves as the noise prediction network $\epsilon_\theta$, predicting the noise $\hat{\bm{\epsilon}}$. The training objective function that ensures the TDSM to learn robust discriminative power is discussed in Sec.~\ref{sec:33}. Finally, Sec.~\ref{sec:34} explains the strategy used during the inference phase to predict the correct label ${\hat{y}}^{u}$ for unseen actions.

\subsection{Embedding Skeleton and Prompt Input}
\label{sec:32}

Following the LDMs \cite{SD, SD3}, we perform the diffusion process in a compact latent space by projecting both skeleton data and prompt into their respective feature spaces. For skeleton data, we adopt GCNs as the architecture for the skeleton encoder $\mathcal{E}_{x}$ that is trained on $\mathcal{D}_{\text{train}}$ using a cross-entropy loss:
\begin{equation}
    \mathcal{L}_{\text{CE}} = - \sum_{k=1}^{|\mathcal{Y}|} \mathbf{y}(k) \log \hat{\mathbf{y}}(k),
\end{equation}
where $\hat{\mathbf{y}} = \mathsf{MLP}(\mathcal{E}_{x}(\mathbf{X}))$ is the predicted class label for a skeleton input $\mathbf{X}_{i}$, $|\mathcal{Y}|$ is the number of seen classes and $\mathbf{y}$ is the one-hot vector of the ground truth label $y$. Once trained, the parameters of skeleton encoder $\mathcal{E}_{x}$ are frozen and used to generate the skeleton latent space representation $\mathbf{z}_{x} = \mathcal{E}_{x}(\mathbf{X})$. After reshaping the feature for the attention layer, $\mathbf{z}_{x}$ is represented in $\mathbb{R}^{M_{x} \times C}$, where $M_{x}$ is the number of skeleton tokens, and $C$ is the feature dimension.

For text encoder $\mathcal{E}_{d}$, we leverage the text prompts to capture rich semantic information about the action labels. Each ground truth (GT) label $y_{p}=y$ is associated with a prompt $\mathbf{d}_{p}$, while a randomly selected wrong label (negative sample) $y_{n} \in \mathcal{Y} \setminus \{y_{p}\}$ is assigned a prompt $\mathbf{d}_{n}$. To encode these prompts, we utilize a pretrained text encoder, such as CLIP \cite{CLIP, openclip}, which provides two types of output features: a global text feature $\mathbf{z}_{g}$ and a local text feature $\mathbf{z}_{l}$. The text encoder's output for a given prompt $\textbf{d}$ can be expressed as:

\begin{equation}
    \left[\mathbf{z}_{g}\mid\mathbf{z}_{l} \right] = \mathcal{E}_d(\textbf{d}),
\end{equation}
where $\left[\;\cdot\mid\cdot\;\right]$ indicates token-wise concatenation, $\mathbf{z}_{g} \in \mathbb{R}^{1 \times C}$ is a global text feature, and $\mathbf{z}_{l} \in \mathbb{R}^{M_{l} \times C}$ is a local text feature, with $M_{l}$ text tokens. For each GT label (positive sample), the text encoder $\mathcal{E}_{d}$ extracts both the global and local text features, denoted as $\mathbf{z}_{g,p}$ and $\mathbf{z}_{l,p}$, respectively. Similarly, for each wrong label (negative sample), the encoder extracts the features $\mathbf{z}_{g,n}$ and $\mathbf{z}_{l,n}$. These four features later guide the diffusion process by conditioning the denoising of noisy skeleton features.

\subsection{Diffusion Process}
\label{sec:33}

Our framework leverages a conditional denoising diffusion process, not to generate data but to learn a discriminative skeleton latent space by fusing skeleton features with text prompts through the reverse diffusion process. Our TDSM is trained to denoise skeleton features such that the resulting latent space becomes discriminative with respect to action labels. Guided by our triplet diffusion (TD) loss, the denoising process conditions on text prompts to strengthen the discriminative fusion of skeleton features and their corresponding prompts. The TD loss encourages correct skeleton-text pairs to be pulled closer in the fused skeleton-text latent space while pushing apart incorrect pairs, enhancing the model's discriminative power.\\
\textbf{Forward process.}  Random Gaussian noise is added to the skeleton feature $\mathbf{z}_{x}$ at a random timestep $t \sim \mathcal{U}(T)$ within total $T$ steps. At each randomly selected step $t$, the noisy feature $\mathbf{z}_{x,t}$ is generated as:
\begin{equation}
    \mathbf{z}_{x,t} = \sqrt{\bar{\alpha}_t} \mathbf{z}_x + \sqrt{1 - \bar{\alpha}_t} \, \bm{\epsilon}.
\end{equation}
\noindent\textbf{Reverse process.}  The Diffusion Transformer $\mathcal{T}_{\text{diff}}$ predicts noise $\hat{\bm{\epsilon}}$ from noisy feature $\mathbf{z}_{x,t}$, conditioned on the global and local text features $\mathbf{z}_{g}$ and $\mathbf{z}_{l}$ at given timestep $t$:
\begin{equation}
    \hat{\bm{\epsilon}} = \mathcal{T}_{\text{diff}}\left(\mathbf{z}_{x,t}, t;\mathbf{z}_{g}, \mathbf{z}_{l}\right).
\end{equation}
Using the shared weights in $\mathcal{T}_{\text{diff}}$, we predict $\hat{\bm{\epsilon}}_{p}$ for positive features $\left(\mathbf{z}_{g,p}, \mathbf{z}_{l,p}\right)$ and $\hat{\bm{\epsilon}}_{n}$ for negative features $\left(\mathbf{z}_{g,n}, \mathbf{z}_{l,n}\right)$. $\mathcal{T}_\text{diff}$ builds upon DiT \cite{DiT}, which has been well-validated for cross-modality alignment in image-text tasks. We adopt it to the skeleton-text domain by: (i) reducing the number of blocks/channels to accommodate the relatively small-scale skeleton data, and (ii) incorporating both global and local text embeddings to enhance skeleton-text alignment. $\mathcal{T}_{\text{diff}}$ is very detailed in the \textit{Supplementary Material}.

\noindent\textbf{Triplet diffusion (TD) loss.} The overall training objective combines a diffusion loss and our reformulated TD loss, which is inspired by the conventional triplet loss \cite{Triplet}, to promote both effective noise prediction and discriminative alignment (fusion). The total loss is defined as:
\begin{equation}
    \mathcal{L}_{\text{total}} = \mathcal{L}_{\text{diff}} + \lambda \mathcal{L}_{\text{TD}},
\end{equation}
where $\mathcal{L}_{\text{diff}}$ ensures accurate denoising, and $\mathcal{L}_{\text{TD}}$ enhances the ability to differentiate between correct and incorrect label predictions. The diffusion loss $\mathcal{L}_{\text{diff}}$ is given by:
\begin{equation}
    \mathcal{L}_{\text{diff}} = \| \bm{\epsilon} - \hat{\bm{\epsilon}}_{p} \|_{2},
\end{equation}
where $\bm{\epsilon}$ is a true noise, and $\bm{\hat{\epsilon}}_{p}$ is a predicted noise for the GT text feature. Our triplet diffusion loss $\mathcal{L}_{\text{TD}}$ is defined as:
\begin{equation}
    \mathcal{L}_{\text{TD}} = \max\left( \| \bm{\epsilon} - \hat{\bm{\epsilon}}_{p} \|_{2} - \| \bm{\epsilon} - \hat{\bm{\epsilon}}_{n} \|_{2} + \tau, \;0 \right),
\end{equation}
where $\hat{\bm{\epsilon}}_{n}$ is the predicted noise for an incorrect (negative) text feature, and $\tau$ is a margin parameter. $\mathcal{L}_{\text{TD}}$ is simple but very effective, encouraging the model to minimize the distance ($\| \bm{\epsilon} - \hat{\bm{\epsilon}}_{p} \|_{2}$) between true noise $\bm{\epsilon}$ and GT prediction $\hat{\bm{\epsilon}}_{p}$ while maximizing the distance ($\| \bm{\epsilon} - \hat{\bm{\epsilon}}_{n} \|_{2}$) for negative predictions $\hat{\bm{\epsilon}}_{n}$, which can ensure discriminative fusion of two modalities in the learned skeleton-text latent space.

\subsection{Inference Phase}
\label{sec:34}
Our approach enhances discriminative fusion through the TD loss, which is designed to denoise GT skeleton-text pairs effectively while preventing the fusion of incorrect pairs within the seen dataset. This selective denoising process promotes a robust fusion of skeleton and text features, allowing the model to develop a discriminative feature space that can generalize to unseen action labels.

\begin{figure}[tbp]
  \centering
  \includegraphics[width=1.0\columnwidth]{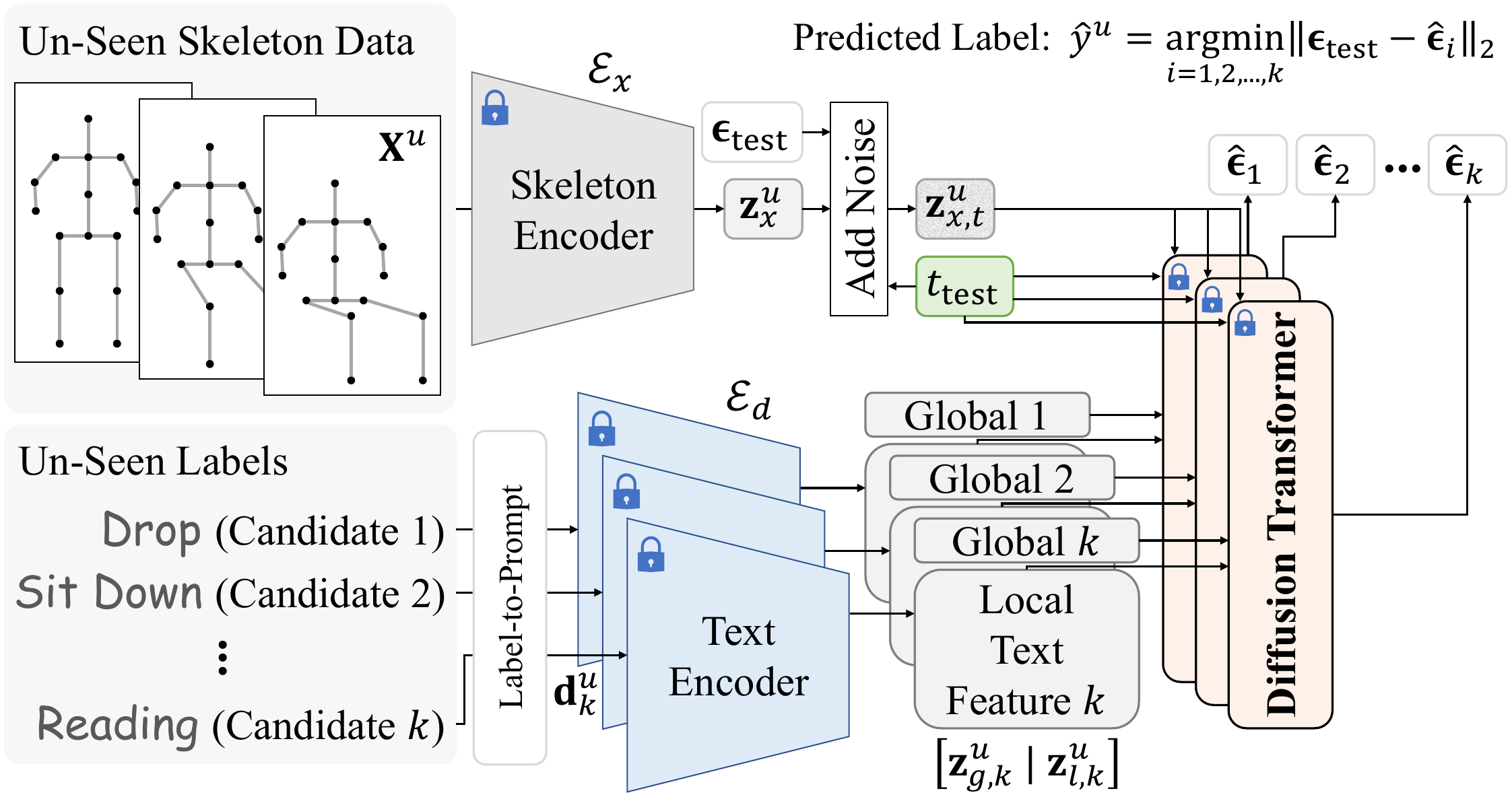}
  \vspace{-0.5cm}
  \caption{Inference framework of our TDSM for ZSAR.}
  \label{fig:infer}
  \vspace{-0.5cm}
\end{figure}

In inference (Fig.~\ref{fig:infer}), each unseen skeleton sequence $\mathbf{X}^{u}$ and its all candidate text prompts are inputted to our TDSM, and the resulting noises for the all candidate text prompts are compared with a fixed GT noise $\bm{\epsilon}_\text{test} \sim \mathcal{N}(\mathbf{0}, \mathbf{I})$. $\mathbf{X}^{u}$ is first encoded into the skeleton latent space through $\mathcal{E}_{x}$ as:
\begin{equation}
    \mathbf{z}_{x}^{u} = \mathcal{E}_{x}(\mathbf{X}^{u}).
\end{equation}
Each candidate action label $y_{k}^{u} \in \mathcal{Y}_u$ is associated with a prompt $\mathbf{d}_{k}^{u}$ that is processed through $\mathcal{E}_{d}$ to extract the global and local text features:
\begin{equation}
    \left[\mathbf{z}_{g,k}^{u}\mid\mathbf{z}_{l,k}^{u}\right]=\mathcal{E}_{d}(\mathbf{d}_{k}^{u}).
\end{equation}
Next, the forward process is performed using a fixed Gaussian noise $\bm{\epsilon}_{\text{test}}$ and a fixed timestep $t_{\text{test}}$ to generate the noisy skeleton feature as:
\begin{equation}
    \mathbf{z}_{x,t}^{u} = \sqrt{\bar{\alpha}_{t_{\text{test}}}} \, \mathbf{z}_{x}^{u} + \sqrt{1 - \bar{\alpha}_{t_{\text{test}}}} \, \bm{\epsilon}_{\text{test}}.
\end{equation}
For $y_{k}^{u}$, the Diffusion Transformer $\mathcal{T}_{\text{diff}}$ predicts noise $\hat{\bm{\epsilon}}_{k}$ as:
\begin{equation}
    \hat{\bm{\epsilon}}_{k} = \mathcal{T}_{\text{diff}}(\mathbf{z}_{x,t}^{u}, t_{\text{test}};\mathbf{z}_{g,k}^{u}, \mathbf{z}_{l,k}^{u}).
\end{equation}
The score for $y_{k}^{u}$ is computed as the $\ell_{2}$-norm between $\bm{\epsilon}_{\text{test}}$ and $\hat{\bm{\epsilon}}_{k}$. The predicted label ${\hat{y}}^{u}$ is then the one that minimizes this distance:
\begin{equation}
    {\hat{y}}^{u} = \arg\min_{k} \| \bm{\epsilon}_{\text{test}} - \hat{\bm{\epsilon}}_{k} \|_{2}.
\end{equation}
This process ensures that the model selects the action label whose text prompt well aligns with the skeleton sequence, enabling accurate zero-shot action recognition for unseen skeleton sequences with unseen action labels. Unlike the generative models that iteratively refine samples, our TDSM performs a \textit{one-step inference} at a fixed timestep, making it efficient and well-suited for discriminative skeleton-text alignment.

%% file: sec/5_experiment.tex
\section{Experiments}
\label{sec:experiment}

\subsection{Datasets}

\noindent\textbf{NTU RGB+D \cite{Ntu60}.} NTU RGB+D dataset, referred to as NTU-60, is one of the largest benchmarks for human action recognition, consisting of 56,880 action samples across 60 action classes. It captures 3D skeleton data, depth maps, and RGB videos using Kinect sensors \cite{Kinetics}, making it a standard for evaluating single- and multi-view action recognition models. The dataset provides a cross-subject split (X-sub) with 40 subjects in total, where 20 subjects are used for training and the remaining 20 for testing. In our experiments, we construct $\mathcal{D}_{\text{train}}$ from the X-sub training set with seen labels and $\mathcal{D}_{\text{test}}$ from the X-sub test set with unseen labels, ensuring a fully zero-shot action recognition setting.

\noindent\textbf{NTU RGB+D 120 \cite{Ntu120}.} NTU RGB+D 120 dataset, referred to as NTU-120, extends the original NTU-60 dataset by adding 60 additional action classes, resulting in a total of 120 classes and 114,480 video samples. The X-sub setting for NTU-120 includes 106 subjects, with 53 used for training and the remaining 53 for testing. We apply the same experimental protocol as NTU-60, using the training set to form $\mathcal{D}_{\text{train}}$ with seen labels and the test set to construct $\mathcal{D}_{\text{test}}$ with unseen labels.

\noindent\textbf{PKU-MMD \cite{PKUMMD}.} PKU-MMD dataset is a large-scale dataset designed for multi-modality action recognition, offering 3D skeleton data and RGB+D recordings. The dataset contains a total of 66 subjects, where 57 subjects are used for training and the remaining 9 for testing. We follow the cross-subject setting to evaluate the generalization of our framework, constructing $\mathcal{D}_{\text{train}}$ with seen labels and $\mathcal{D}_{\text{test}}$ with unseen labels.

\begin{table*}[tbp]
    \scriptsize
    \centering
    \resizebox{1.0\textwidth}{!}{
    \def\arraystretch{1.2}
    \begin{tabular} {l|l|c|c|c|c|c|c|c|c}
        \Xhline{2\arrayrulewidth}
        \multirow{2}{*}{Methods} & \multirow{2}{*}{Publications} &\multicolumn{4}{c|}{NTU-60\;(Acc, \%)} & \multicolumn{4}{c}{NTU-120 \;(Acc, \%)} \\
        \cline{3-10}
        & & 55/5 split & 48/12 split & 40/20 split & 30/30 split & 110/10 split & 96/24 split & 80/40 split & 60/60 split \\
        \hline
        ReViSE \cite{ReViSE} & ICCV 2017 & 53.91 & 17.49 & 24.26 & 14.81 & 55.04 & 32.38 & 19.47 & 8.27 \\
        JPoSE \cite{JPoSE} & ICCV 2019 & 64.82 & 28.75 & 20.05 & 12.39 & 51.93 & 32.44 & 13.71 & 7.65 \\
        CADA-VAE \cite{CADAVAE} & CVPR 2019 & 76.84 & 28.96 & 16.21 & 11.51 & 59.53 & 35.77 & 10.55 & 5.67 \\
        SynSE \cite{SynSE} & ICIP 2021 & 75.81 & 33.30 & 19.85 & 12.00 & 62.69 & 38.70 & 13.64 & 7.73 \\
        SMIE \cite{SMIE} & ACM MM 2023 & 77.98 & 40.18 & - & - & 65.74 & 45.30 & - & - \\
        PURLS \cite{PURLS} & CVPR 2024 & 79.23 & 40.99 & \color{blue}{\underline{31.05}} & \color{blue}{\underline{23.52}} & \color{blue}{\underline{71.95}} & \color{blue}{\underline{52.01}} & \color{blue}{\underline{28.38}} & \color{blue}{\underline{19.63}} \\
        SA-DVAE \cite{SADVAE} & ECCV 2024 & \color{blue}{\underline{82.37}} & 41.38 & - & - & 68.77 & 46.12 & - & - \\
        STAR \cite{STAR} & ACM MM 2024 & 81.40 & \color{blue}{\underline{45.10}} & - & - & 63.30 & 44.30 & - & - \\
        \textbf{TDSM (Ours)} & \textbf{-} & \textbf{\color{red}{86.49}} & \textbf{\color{red}{56.03}} & \textbf{\color{red}{36.09}} & \textbf{\color{red}{25.88}} & \textbf{\color{red}{74.15}} & \textbf{\color{red}{65.06}} & \textbf{\color{red}{36.95}} & \textbf{\color{red}{27.21}} \\
        \Xhline{2\arrayrulewidth}
    \end{tabular}}
    \vspace{-0.2cm}
    \caption{Top-1 accuracy results of various zero-shot skeleton-based action recognition (ZSAR) methods evaluated on the SynSE and PURLS benchmarks for the NTU-60 and NTU-120 datasets. Each split is denoted as X/Y, where X represents the number of seen classes and Y the number of unseen classes. The results in \textbf{\color{red}{red}} highlight the best-performing model, while those in {\color{blue}{\underline{blue}}} indicate the second-best. For our TDSM framework, the reported accuracy is the average value obtained from 10 trials, each with different Gaussian noise.}
    \vspace{-0.3cm}
  \label{tab:main}
\end{table*}

\subsection{Experiment Setup}
\begin{figure*}[tbp]
  \centering
  \includegraphics[width=0.99\textwidth]{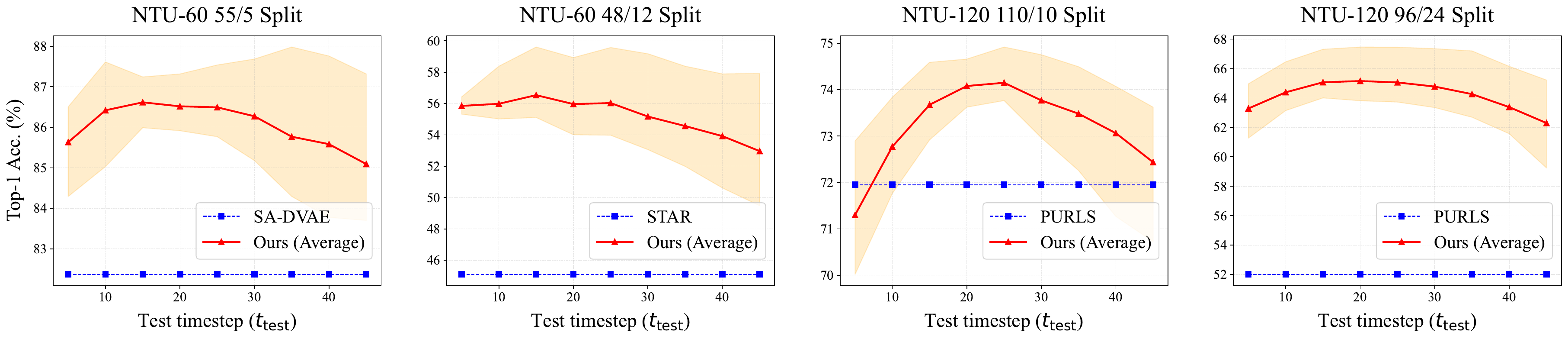}
  \vspace{-0.2cm}
  \caption{Effect of varying inference timesteps $t_{\text{test}}$ across multiple datasets. Each plot shows the top-1 accuracy trend on the NTU-60 and NTU-120 datasets under different splits. The solid {\color{red}{red line}} represents the average accuracy of our method, with the shaded {\color{orange}{orange area}} indicating the variation in accuracy across 10 different random Gaussian noise instances. Dashed {\color{blue}{blue line}} corresponds to the second-best method in each benchmark.}
  \vspace{-0.3cm}
  \label{fig:graph}
\end{figure*}

Our TDSM were implemented in PyTorch \cite{Pytorch} and conducted on a single NVIDIA GeForce RTX 3090 GPU. Our model variants were trained for 50,000 iterations, with a warm-up period of 100 steps. We employed the AdamW optimizer \cite{AdamW} with a learning rate of $1 \times 10^{-4}$ and a weight decay of 0.01. A cosine-annealing scheduler \cite{CosineAnneal} was used to dynamically update the learning rate at each iteration. The batch size was set to 256, but for the TD loss computation, the batch was duplicated for positive and negative samples, resulting in an effective batch size of 512. Through empirical validation, the loss weight $\lambda$ and margin $\tau$ were both set to 1.0. The diffusion process was trained with a total timestep of $T=50$. For inference, the optimal timestep $t_{\text{test}} = 25$ was selected based on accuracy trends across datasets (Fig.~\ref{fig:graph}). For the SynSE \cite{SynSE} and PURLS \cite{PURLS} seen and unseen split settings, we used the Shift-GCN \cite{ShiftGCN} architecture as our skeleton encoder. In the SMIE \cite{SMIE} split setting, we employed the ST-GCN \cite{STGCN} structure. We adopted Shift-GCN and ST-GCN because the majority of previous ZSAR studies use them, ensuring a fair comparison without a confounding factor of potentially more powerful encoders. For fair comparison, we used the same text prompts employed in existing works. Across all tasks, we adopted the text encoder from CLIP \cite{CLIP, openclip} to transform text prompts into latent representations.

\subsection{Performance Evaluation}
\noindent\textbf{Evaluation on SynSE \cite{SynSE} and PURLS \cite{PURLS} benchmarks.}  Table~\ref{tab:main} presents the performance comparison on the SynSE and PURLS benchmark splits across the NTU-60 and NTU-120 datasets. The SynSE benchmark focuses on standard settings, offering 55/5 and 48/12 splits on NTU-60, and 110/10 and 96/24 splits on NTU-120. These settings assess the model’s ability to generalize across typical seen-unseen splits. In contrast, the PURLS benchmark presents more extreme cases with 40/20 and 30/30 splits on NTU-60, and 80/40 and 60/60 splits on NTU-120, introducing higher levels of complexity by increasing the proportion of unseen labels in the test set. To account for the stochastic nature of noise during inference, we averaged the 10 runs with different Gaussian noise realizations. As shown in Table~\ref{tab:main}, our TDSM significantly outperforms the very recent state-of-the-art results across all benchmark splits, demonstrating superior generalization and robustness for various splits. Specifically, TDSM outperforms the existing methods on both standard (SynSE) and extreme (PURLS) settings, with 4.12\%-point, 9.93\%-point, 5.04\%-point, and 2.36\%-point improvements in top-1 accuracy on the NTU-60 55/5, 48/12, 40/20, and 30/30 splits, respectively, compared to the second best models. Also compared to the second best models, our TDSM attains 2.20\%-point, 13.05\%-point, 8.57\%-point, and 7.58\%-point accuracy improvements on the NTU-120 splits, further validating its scalability to larger datasets and more complex unseen classes.

\begin{table}[tbp]
    \scriptsize
    \centering
    \resizebox{1.0\columnwidth}{!}{
    \def\arraystretch{1.2}
    \begin{tabular} {l|c|c|c}
        \Xhline{2\arrayrulewidth}
        \multirow{2}{*}{Methods} & NTU-60\;(Acc, \%) & NTU-120\;(Acc, \%) & PKU-MMD\;(Acc, \%) \\
        \cline{2-4}
        & 55/5 split & 110/10 split & 46/5 split \\
        \hline
        ReViSE \cite{ReViSE} & 60.94 & 44.90 & 59.34 \\
        JPoSE \cite{JPoSE} & 59.44 & 46.69 & 57.17 \\
        CADA-VAE \cite{CADAVAE} & 61.84 & 45.15 & 60.74 \\
        SynSE \cite{SynSE} & 64.19 & 47.28 & 53.85 \\
        SMIE \cite{SMIE} & 65.08 & 46.40 & 60.83 \\
        SA-DVAE \cite{SADVAE} & \color{blue}{\underline{84.20}} & \color{blue}{\underline{50.67}} & 66.54 \\
        STAR \cite{STAR} & 77.50 & - & \color{blue}{\underline{70.60}} \\
        \textbf{TDSM (Ours)} & \textbf{\color{red}{88.88}} & \textbf{\color{red}{69.47}} & \color{red}{\textbf{70.76}} \\
        \Xhline{2\arrayrulewidth}
    \end{tabular}}
    \vspace{-0.2cm}
    \caption{Top-1 accuracy results of various ZSAR methods evaluated on the NTU-60, NTU-120, and PKU-MMD datasets under the SMIE benchmark. The reported values are the average performance across three splits.}
  \label{tab:smie}
\end{table}

\begin{table}[tbp]
    \scriptsize
    \centering
    \resizebox{1.0\columnwidth}{!}{
    \def\arraystretch{1.2}
    \begin{tabular} {c|c|c|c|c|c}
        \Xhline{2\arrayrulewidth}
        \multirow{2}{*}{$\mathcal{L}_{\text{diff}}$} & \multirow{2}{*}{$\mathcal{L}_{\text{TD}}$} &\multicolumn{2}{c|}{NTU-60\;(Acc, \%)} & \multicolumn{2}{c}{NTU-120 \;(Acc, \%)} \\
        \cline{3-6}
        & & 55/5 split & 48/12 split & 110/10 split & 96/24 split \\
        \hline
        \checkmark & & 79.87 & 53.03 & 72.44 & 57.65 \\
        & \checkmark & 80.90 & 54.36 & 70.73 & 60.95 \\
        \checkmark & \checkmark & \textbf{\color{red}{86.49}} & \textbf{\color{red}{56.03}} & \textbf{\color{red}{74.15}} & \textbf{\color{red}{65.06}} \\
        \Xhline{2\arrayrulewidth}
    \end{tabular}}
    \vspace{-0.2cm}
    \caption{Ablation study on loss function configurations. The results compare models trained with only the diffusion loss \( \mathcal{L}_{\text{diff}} \), only the triplet diffusion loss \( \mathcal{L}_{\text{TD}} \), and their combination.}
  \label{tab:loss}
\end{table}

\noindent\textbf{Evaluation on SMIE \cite{SMIE} benchmark.} The SMIE benchmark provides three distinct splits to evaluate the generalization capability of models across different sets of unseen labels. Each split ensures that unseen labels do not overlap with seen ones, thereby rigorously testing the model's ability to recognize new classes without prior exposure. For fair comparison, the reported performance is the average of the three splits. In this benchmark, as shown in Table~\ref{tab:smie}, our TDSM outperforms the other methods, demonstrating strong generalization across all evaluated datasets.

\subsection{Ablation Studies}

\noindent\textbf{Effect of loss function design.} The ablation study shown in Table~\ref{tab:loss} evaluates the impact of combining the diffusion loss $\mathcal{L}_{\text{diff}}$ and the triplet diffusion (TD) loss $\mathcal{L}_{\text{TD}}$ on the model's performance. The results demonstrate that leveraging both losses yields superior performance compared to using either one alone. Specifically, when only $\mathcal{L}_{\text{diff}}$ is employed, the model focuses on accurately denoising the skeleton features conditioned on the text prompts but may lack sufficient discriminative power between similar actions. Conversely, applying only $\mathcal{L}_{\text{TD}}$ enhances discriminative fusion but without ensuring optimal noise prediction. The combination of both losses strikes a balance, ensuring discriminative fusion, which resulting in the highest performance across all evaluated splits. 

\begin{table}[tbp]
    \scriptsize
    \centering
    \resizebox{1.0\columnwidth}{!}{
    \def\arraystretch{1.25}
    \begin{tabular} {c|c|c|c|c|c}
        \Xhline{2\arrayrulewidth}
        \multirow{2}{*}{\makecell{Global\\$\mathbf{z}_g$}} & \multirow{2}{*}{\makecell{Local\\$\mathbf{z}_l$}} &\multicolumn{2}{c|}{NTU-60\;(Acc, \%)} & \multicolumn{2}{c}{NTU-120 \;(Acc, \%)} \\
        \cline{3-6}
        & & 55/5 split & 48/12 split & 110/10 split & 96/24 split \\
        \hline
        \checkmark & & 83.41 & 51.50 & 70.14  & 61.90  \\
        & \checkmark & 83.33 & 52.63 & 69.95 & 62.10 \\
        \checkmark & \checkmark & \textbf{\color{red}{86.49}} & \textbf{\color{red}{56.03}} & \textbf{\color{red}{74.15}} & \textbf{\color{red}{65.06}} \\
        \Xhline{2\arrayrulewidth}
    \end{tabular}}
    \vspace{-0.2cm}
    \caption{Ablation study on text feature types. The results compare models trained with only \( \mathbf{z}_g \), only \( \mathbf{z}_l \), and their combination.}
    \vspace{-0.3cm}
  \label{tab:text}
\end{table}

\noindent\textbf{Contribution of global and local text features.} Our TDSM framework employs two types of text features for skeleton-text matching: a global text feature $\mathbf{z}_g$ that encodes the entire sentence as a single token, and local text features $\mathbf{z}_l$ that preserve token-level details for each word in the sentence. As shown in Table~\ref{tab:text}, combining the global and local text features achieves the best performance, outperforming the models that use either feature independently. The global text feature provides overall discriminative power by capturing high-level semantics of the action description, enabling robust matching across diverse action categories. Meanwhile, the local text features retain finer details that are effective for distinguishing subtle differences between semantically similar actions.

\noindent\textbf{Impact of total timesteps $T$.} The results of the ablation study on diffusion timesteps are shown in Table~\ref{tab:timestep}. We observe that the choice of $T$ has a significant impact on performance across all datasets and splits. When $T$ is too small, the model tends to overfit, as the problem becomes too simple, limiting the diversity in noise added to the skeleton features during training. On the other hand, too large $T$ values introduces diverse noise strengths, making it challenging for the model to denoise effectively, which deteriorates performance. The best $T$ is found empirically with $T=50$, striking a balance between maintaining a challenging task and avoiding overfitting.

\noindent\textbf{Effect of random Gaussian noise.} To examine the role of Gaussian noise in training, we conducted ablation studies by using fixed Gaussian noise during both training and inference, instead of introducing new random noise at each training step. As shown in Table~\ref{tab:fixed}, using fixed Gaussian noise overly simplifies the learning process, causing the network to overfit specific noise patterns and reducing its generalization ability. In contrast, introducing random Gaussian noise at each step increases variability in the learning process, acting as a regularization mechanism that prevents overfitting. This enhances model robustness and improves the alignment between skeleton features and text prompts.

\noindent\textbf{Impact of timestep $t_{\text{test}}$ and noise $\bm{\epsilon}_{\text{test}}$ in inference.} We conducted experiments across different test timesteps $t_{\text{test}} \in [0, 50]$ and observed the accuracy trends on multiple datasets, as illustrated in Fig.~\ref{fig:graph}. Based on these observations, we set $t_{\text{test}} = 25$ for all experiments. To examine the impact of noise during inference, we repeated experiments with 10 different random Gaussian noise samples. In Fig.~\ref{fig:graph}, the shaded orange regions in the graphs depict the variances in top-1 accuracy due to noise, while the red lines represent the average accuracy. The blue dashed lines indicate the second-best method’s accuracy for comparison. Our analysis shows that while noise variations can cause up to a $\pm2.5\%$-point changes in top-1 accuracy at $t_{\text{test}} = 25$, our TDSM shows consistently outperforming the state-of-the-art methods regardless of noise levels.

\begin{table}[tbp]
    \scriptsize
    \centering
    \resizebox{0.9\columnwidth}{!}{
    \def\arraystretch{1.2}
    \begin{tabular} {c|c|c|c|c}
        \Xhline{2\arrayrulewidth}
        \multirow{2}{*}{Total $T$} & \multicolumn{2}{c|}{NTU-60\;(Acc, \%)} & \multicolumn{2}{c}{NTU-120 \;(Acc, \%)} \\
        \cline{2-5}
        & 55/5 split & 48/12 split & 110/10 split & 96/24 split \\
        \hline
        1 & 85.03 & 44.10 & 69.91 & 60.35 \\
        10 & 84.51 & 50.89 & 69.97 & 62.04\\
        50 & \textbf{\color{red}{86.49}} & \textbf{\color{red}{56.03}} & \textbf{\color{red}{74.15}} & \textbf{\color{red}{65.06}} \\
        100 & 83.48 & 56.27 & 71.05 & 64.57 \\
        500 & 81.34 & 53.43 & 71.93 & 60.81 \\
        \Xhline{2\arrayrulewidth}
    \end{tabular}}
    \vspace{-0.2cm}
    \caption{Ablation study on the impact of total timesteps $T$ in the training of the diffusion process.}
  \label{tab:timestep}
\end{table}

\begin{table}[tbp]
    \scriptsize
    \centering
    \resizebox{0.9\columnwidth}{!}{
    \def\arraystretch{1.2}
    \begin{tabular} {c|c|c|c|c}
        \Xhline{2\arrayrulewidth}
        \multirow{2}{*}{\makecell{Gaussian\\noise $\bm\epsilon$}} & \multicolumn{2}{c|}{NTU-60\;(Acc, \%)} & \multicolumn{2}{c}{NTU-120 \;(Acc, \%)} \\
        \cline{2-5}
        & 55/5 split & 48/12 split & 110/10 split & 96/24 split \\
        \hline
        Fixed & 76.40 & 44.25 & 64.01 & 52.21 \\
        Random & \textbf{\color{red}{86.49}} & \textbf{\color{red}{56.03}} & \textbf{\color{red}{74.15}} & \textbf{\color{red}{65.06}} \\
        \Xhline{2\arrayrulewidth}
    \end{tabular}}
    \vspace{-0.2cm}
    \caption{Ablation study on the effect of noise $\bm{\epsilon}$ during training.}
    \vspace{-0.3cm}
  \label{tab:fixed}
\end{table}

%% file: sec/6_conclusion.tex
\section{Conclusion}
\label{sec:conclusion}
Our TDSM is the first framework to apply diffusion models to zero-shot skeleton-based action recognition. The selective denoising process promotes a robust fusion of skeleton and text features, allowing the model to develop a discriminative feature space that can generalize to unseen action labels. Also, our approach enhances discriminative fusion through the TD loss which is designed to denoise GT skeleton-text pairs effectively while preventing the fusion of incorrect pairs within the seen dataset. Extensive experiments show that our TDSM significantly outperforms the very recent SOTA models with large margins for various benchmark datasets.

%% file: sec/X_suppl.tex
\clearpage
\maketitlesupplementary
\appendix

\section{Limitations}

\subsection{Sensitivity to Noise Variation}  

As illustrated in Fig.~\ref{fig:graph}, although our method achieves superior performance, its results exhibit some sensitivity to noise $\bm{\epsilon}_\text{test}$ during inference. Recent studies \cite{Lotus} have suggested that predicting the initial state $\mathbf{z}_{x}$ during the reverse diffusion process yields more stable results compared to direct noise ($\bm{\epsilon}$) prediction, especially under varying noise conditions. As part of future work, we plan to explore this refinement to enhance the robustness against noise fluctuations.

\begin{figure*}[htbp]
  \centering
  \includegraphics[width=1.0\textwidth]{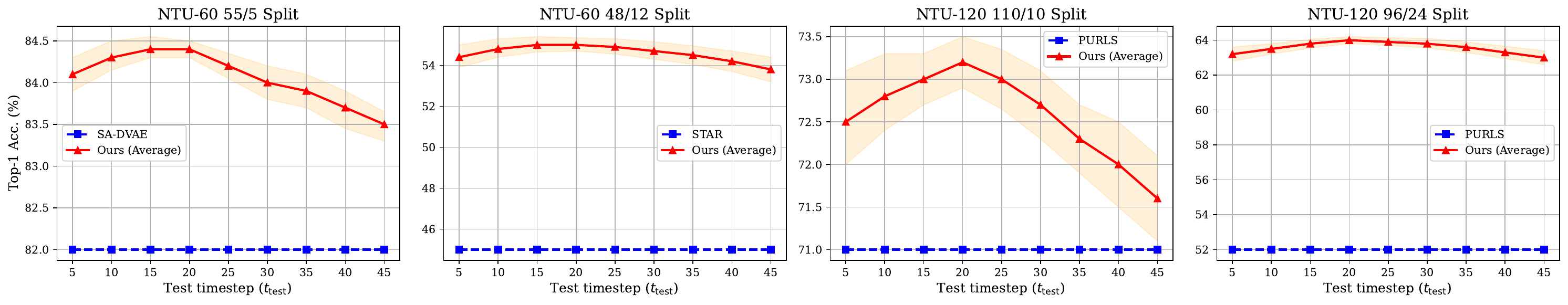}
  \vspace{-0.2cm}
  \caption{Effect of varying inference timesteps $t_{\text{test}}$ across multiple datasets. Each plot shows the top-1 accuracy trend on the NTU-60 and NTU-120 datasets under different splits. The solid {\color{red}{red line}} represents the average accuracy of our method, with the shaded {\color{orange}{orange area}} indicating the variation in accuracy across 10 different random Gaussian noise instances. Dashed {\color{blue}{blue line}} corresponds to the second-best method in each benchmark.}
  \label{fig:supple_graph}
\end{figure*}

\noindent \textbf{Predicting $\mathbf{z}_x$.} Our model exhibits somewhat sensitivity to noise during inference. To address this, we additionally experimented with predicting $\mathbf{z}_x$ instead of noise $\bm{\epsilon}$. As shown in Fig.~\ref{fig:supple_graph}, noise-induced fluctuation is reduced by 5$\times$, with minimal performance drop and SOTA-level accuracy maintained.

\section{Additional Discussions on Results}

\subsection{Results on More Complex Datasets}

We provide additional results on Kinetics-200 and Kinetics-400 datasets \cite{Kinetics} in Table~\ref{tab:kinetics1} and Table~\ref{tab:kinetics2}. For consistency, we use the same skeleton encoder and same text prompts as PURLS \cite{PURLS}. Notably, despite using only a \textit{single} text prompt per action, our method achieves state-of-the-art performance across all data splits, outperforming prior approaches that leverage multiple text prompts.

\subsection{More Comparison with BSZSL}

We did additional comparison with BSZSL \cite{liu2025beyond} that utilizes both text and RGB modalities. Without relying on RGB input, our TDSM in the Table~\ref{tab:bszsl} outperforms BSZSL in 3 out of 4 splits across NTU-60 and NTU-120 datasets.

\subsection{Analysis of Split Settings}

\begin{figure*}[tbp]
  \centering
  \includegraphics[width=0.67\textwidth]{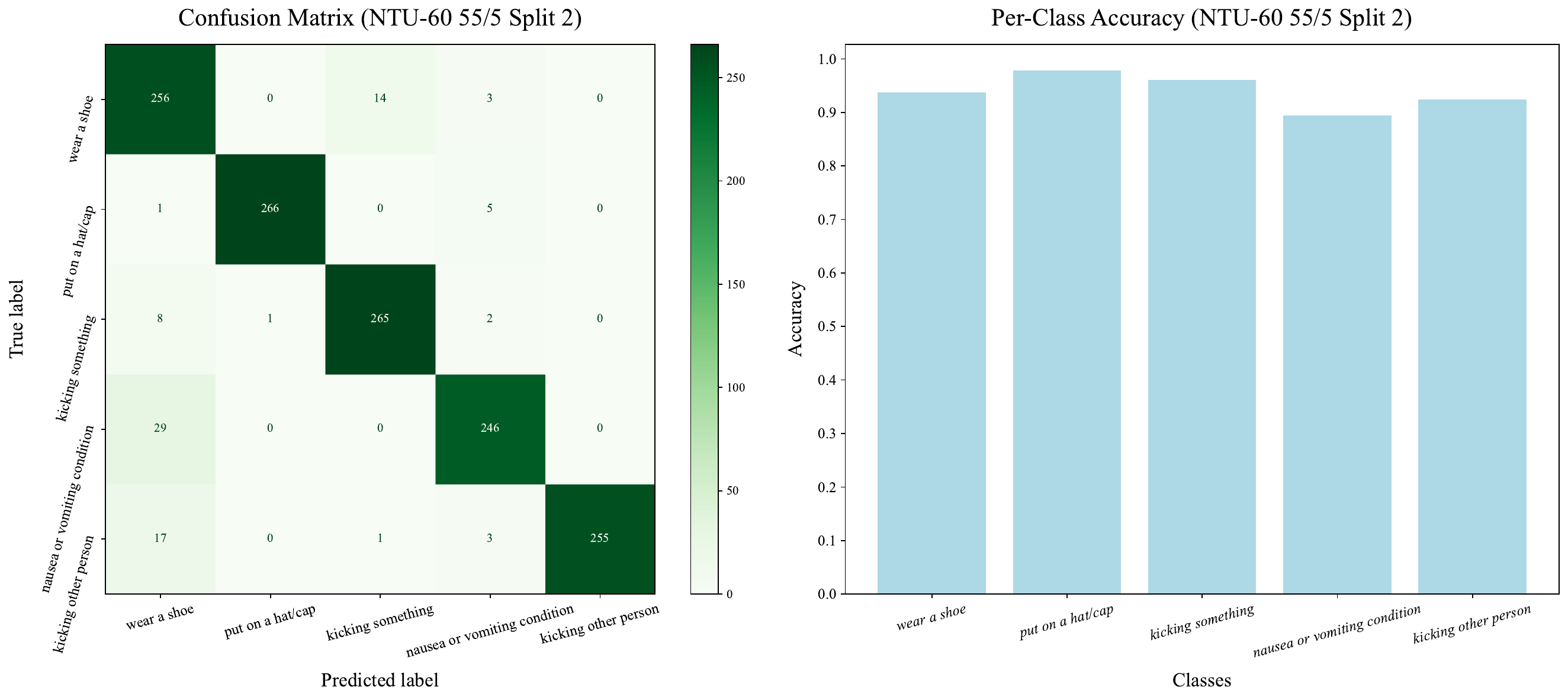}
  \vspace{-0.2cm}
  \caption{Confusion matrix and per-class top-1 accuracy visualization for NTU-60 55/5 Split 2.}
  \label{fig:vis1}
\end{figure*}

Table~\ref{tab:smie} presents the average performance of our model across split 1, split 2, and split 3 on the SMIE \cite{SMIE} benchmark. For a more detailed analysis, Table~\ref{tab:smie_detail} reports the performance for each individual split. Notably, in the NTU-60 55/5 split, our TDSM achieves the highest performance for split 2. The unseen classes in this split---``\textit{wear a shoe}'', ``\textit{put on a hat/cap}'', ``\textit{kicking something}'', ``\textit{nausea or vomiting condition}'', and ``\textit{kicking other person}''---exhibit clear and distinct motion patterns. For example, ``\textit{wear a shoe}'' involves downward torso motion, ``\textit{put on a hat/cap}'' features upward hand movements, ``\textit{kicking something}'' emphasizes significant leg activity, ``\textit{nausea or vomiting condition}'' depicts upper body contraction, and ``\textit{kicking other person}'' is unique as it involves two skeletons interacting. These distinct characteristics make our TDSM easier to distinguish the classes, leading to higher performance. Fig.~\ref{fig:vis1} illustrates this trend through the confusion matrix and per-class accuracy visualization, highlighting the clear separability of these actions.

\begin{figure*}[tbp]
  \centering
  \includegraphics[width=0.67\textwidth]{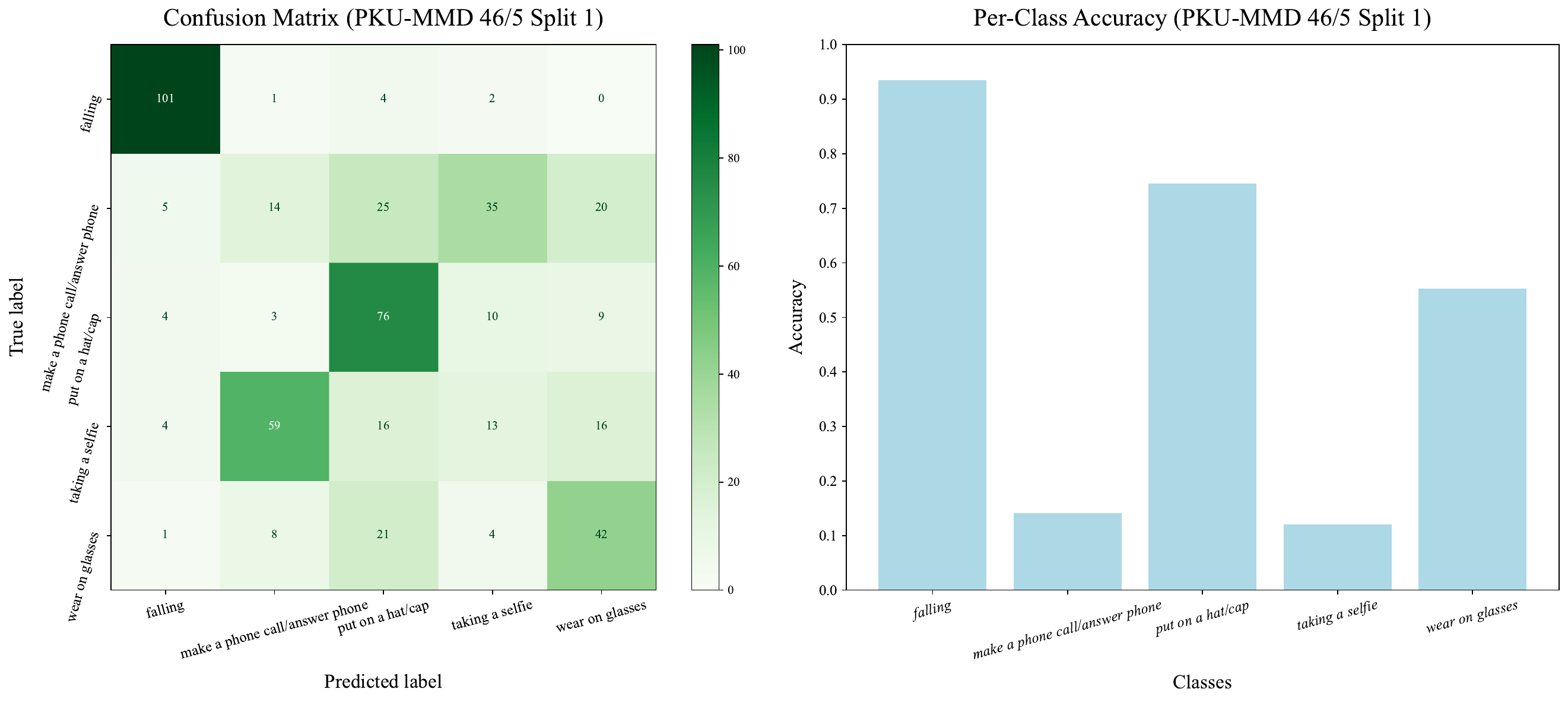}
  \vspace{-0.2cm}
  \caption{Confusion matrix and per-class top-1 accuracy visualization for PKU-MMD 46/5 Split 1.}
  \label{fig:vis2}
\end{figure*}

In contrast, our TDSM shows relatively lower performance for the split 1 in the PKU-MMD 46/5 split, although the split 1 contains fewer unseen classes (``\textit{falling}'', ``\textit{make a phone call/answer phone}'', ``\textit{put on a hat/cap}'', ``\textit{taking a selfie}'', and ``\textit{wear on glasses}''). Except for ``\textit{falling}'', the remaining classes involve similar upward hand movements and interactions with objects (e.g., phones, hats, glasses) that are not explicitly visible in skeleton data. This lack of contextual information makes it significantly harder to distinguish these actions, resulting in degraded performance. As visualized in Fig.~\ref{fig:vis2}, the confusion matrix and per-class accuracy further reveal the challenge of separating actions with overlapping motion patterns, emphasizing the limitations of skeleton-only data when distinguishing semantically similar actions. These observations underscore the importance of distinct motion patterns in unseen classes for robust zero-shot recognition.

\subsection{Potential Training-Inference Mismatch}

Our TDSM adopts an one-step inference framework, where both training and inference are consistently performed with the same total number of timesteps $T$. Empirically, we found that performing one-step inference at $t_{\text{test}} = T/2$ (e.g., $t_{\text{test}} = 50$ when $T = 100$) provides the best trade-off between noise and structure. So, no distributional mismatch exists b/w training and inference in our setting.

\subsection{More Explanation about Text Feature}

Compared to PURLS \cite{PURLS} where local textual features are obtained from six separate body-part-specific descriptions, our TDSM extracts both global and local features from a single unified sentence, yielding and $\mathbf{z}_g$ and $\mathbf{z}_l$ which are described in details.

\begin{table}[tbp]
    \scriptsize
    \centering
    \resizebox{1.0\columnwidth}{!}{
    \def\arraystretch{1.2}
    \begin{tabular} {l|c|c|c|c}
        \Xhline{2\arrayrulewidth}
        \multirow{2}{*}{Methods} &\multicolumn{4}{c}{Kinetics-200\;(Acc, \%)} \\
        \cline{2-5}
        & 180/20 split & 160/40 split & 140/60 split & 120/80 split \\
        \hline
        ReViSE \cite{ReViSE} & 24.95 & 13.28 & 8.14 & 6.23 \\
        DeViSE \cite{DeViSE} & 22.22 & 12.32 & 7.97 & 5.65 \\
        PURLS \cite{PURLS} (1 text) & 25.96 & 15.85 & 10.23 & 7.77 \\
        PURLS \cite{PURLS} (7 text) & 32.22 & 22.56 & 12.01 & 11.75 \\
        \hline
        \textbf{TDSM} (1 text) & \textbf{\color{red}{38.18}} & \textbf{\color{red}{24.43}} & \textbf{\color{red}{15.28}} & \color{red}{\textbf{13.09}} \\
        \Xhline{2\arrayrulewidth}
    \end{tabular}}
    \vspace{-0.2cm}
    \caption{Top-1 accuracy results of TDSM evaluated on the Kinetics-200 dataset under the PURLS \cite{PURLS} benchmark.}
    \label{tab:kinetics1}
\end{table}

\begin{table}[tbp]
    \scriptsize
    \centering
    \resizebox{1.0\columnwidth}{!}{
    \def\arraystretch{1.2}
    \begin{tabular} {l|c|c|c|c}
        \Xhline{2\arrayrulewidth}
        \multirow{2}{*}{Methods} &\multicolumn{4}{c}{Kinetics-400\;(Acc, \%)} \\
        \cline{2-5}
        & 360/40 split & 320/80 split & 300/100 split & 280/120 split \\
        \hline
        ReViSE \cite{ReViSE} & 20.84 & 11.82 & 9.49 & 8.23 \\
        DeViSE \cite{DeViSE} & 18.37 & 10.23 & 9.47 & 8.34 \\
        PURLS \cite{PURLS} (1 text) & 22.50 & 15.08 & 11.44 & 11.03 \\
        PURLS \cite{PURLS} (7 text) & 34.51 & 24.32 & 16.99 & 14.28 \\
        \hline
        \textbf{TDSM} (1 text) & \textbf{\color{red}{38.92}} & \textbf{\color{red}{26.24}} & \textbf{\color{red}{18.45}} & \color{red}{\textbf{16.10}} \\
        \Xhline{2\arrayrulewidth}
    \end{tabular}}
    \vspace{-0.2cm}
    \caption{Top-1 accuracy results of TDSM evaluated on the Kinetics-400 dataset under the PURLS \cite{PURLS} benchmark.}
    \label{tab:kinetics2}
\end{table}

\begin{table}[tbp]
    \scriptsize
    \centering
    \resizebox{1.0\columnwidth}{!}{
    \def\arraystretch{1.2}
    \begin{tabular} {l|c|c|c|c|c|c}
        \Xhline{2\arrayrulewidth}
        \multirow{2}{*}{Methods} & \multicolumn{2}{c|}{Modality} & \multicolumn{2}{c|}{NTU-60\;(Acc, \%)} & \multicolumn{2}{c}{NTU-120 \;(Acc, \%)} \\
        \cline{2-7}
        & Text & RGB & 55/5 split & 48/12 split & 110/10 split & 96/24 split \\
        \hline
        BSZSL \cite{liu2025beyond} & \checkmark & \checkmark & 83.04 & 52.96 & \textbf{\color{red}{77.69}} & 56.12 \\
        \textbf{TDSM} & \checkmark & & \textbf{\color{red}{86.49}} & \textbf{\color{red}{56.03}} & 74.15 & \textbf{\color{red}{65.06}} \\
        \Xhline{2\arrayrulewidth}
    \end{tabular}}
    \vspace{-0.2cm}
    \caption{Top-1 accuracy results of BSZSL evaluated on the SynSE and PURLS benchmarks for the NTU-60 and NTU-120 datasets.}
    \label{tab:bszsl}
\end{table}

\begin{table}[tbp]
    \scriptsize
    \centering
    \resizebox{1.0\columnwidth}{!}{
    \def\arraystretch{1.25}
    \begin{tabular} {l|c|c|c}
        \Xhline{2\arrayrulewidth}
        \multirow{2}{*}{\makecell{\textbf{TDSM}\\ \textbf{(Ours)}}} & NTU-60\;(Acc, \%) & NTU-120\;(Acc, \%) & PKU-MMD\;(Acc, \%) \\
        \cline{2-4}
        & 55/5 split & 110/10 split & 46/5 split \\
        \hline
        Split 1 & 87.97 & 74.45 & 57.40 (Fig.~\ref{fig:vis2}) \\
        Split 2 & 96.06 (Fig.~\ref{fig:vis1}) & 63.91 & 76.92 \\
        Split 3 & 82.60 & 70.04 & 77.97 \\
        \hline
        Average & 88.88 & 69.47 & 70.76 \\
        \Xhline{2\arrayrulewidth}
    \end{tabular}}
    \vspace{-0.2cm}
    \caption{Top-1 accuracy results of our TDSM evaluated on the NTU-60, NTU-120, and PKU-MMD datasets under the SMIE \cite{SMIE} benchmark.}
    \label{tab:smie_detail}
\end{table}

\section{Discussion on ZSAR Methods}

The key contribution of our work lies not in each individual component (e.g., DiT architecture, loss function but in proposing a new framework for ZSAR that effectively bridges the cross-modality gap between skeleton and text. Previous VAE-based or contrastive learning(CL)-based methods attempt \textit{direct} alignment between skeleton and text latents, but the inherent large modality gap limits their effectiveness (Sec.~\ref{sec:22}). To address this, we use a diffusion process---already shown to be powerful in image-text alignment---and adapt it for a discriminative zero-shot action recognition task. The implication of our DM-based TDSM is very meaningful as Table~\ref{tab:novelty}, which has brought out significant performance improvement with 2.36 to 13.05\%-point.

\subsection{Difference against Previous ZSAR Methods}

As illustrated in Table~\ref{tab:novelty}, previous VAE-based and CL-based methods rely on \textit{explicit point-wise alignment}, minimizing cross-reconstruction error or feature distances directly between skeleton and text features. But, our TDSM aligns the two modalities implicitly by learning to denoise a noisy skeleton feature in a single reverse diffusion step conditioned on a text feature.

\subsection{Why Diffusion is Effective for ZSAR?}

Diffusion models are known for their strong cross-modal alignment capabilities, enabled through conditioning mechanisms that integrate signals. Our TDSM leverages this property using an one-step reverse diffusion process conditioned on a text embedding, to denoise a skeleton feature. We believed this property of diffusion (its ability to integrate semantic guidance during denoising) would be particularly effective for ZSAR, where bridging modality gaps is critical. To the best of our knowledge, our TDSM is the first to apply diffusion in this discriminative alignment setting for ZSAR, validating its effectiveness across multiple benchmarks.

In comparison with the previous work \cite{chen2023executing, huang2024stablemofusion, gong2023diffpose, holmquist2023diffpose, wei2025learning, foo2024action}, they are based on diffusion models and have generation tasks while our TDSM utilizes the property of diffusion model's strong cross-modality alignment for discriminative tasks, not for generation tasks. Note that \cite{foo2024action} predicts actions by generating visual representations in an iterative sampling process, while our TDSM utilizes a diffusion model in a single-step inference without generating any feature for action classification.

\begin{table*}[htbp]
    \scriptsize
    \centering
    \resizebox{1.0\textwidth}{!}{
    \def\arraystretch{1.1}
    \begin{tabular}{l|l|c}
        \Xhline{2\arrayrulewidth}
        Methods & Characteristics & Limitations \\ 
        \hline
        \multirow{2}{*}{VAE-based} & Reconstructs skeleton-text feature pairs via \textbf{cross-reconstruction,} & \multirow{4}{*}{\makecell{Modality gap due to direct alignment}}\\
        & recovering skeleton features from text and vice versa & \\
        \cline{1-2}
        \multirow{2}{*}{CL-based}  & Aligns skeleton and text features by \textbf{minimizing feature distance} & \\
        & through contrastive learning & \\
        \hline
        \multirow{3}{*}{\makecell{\textbf{TDSM} \\ \textbf{(Ours)}}} & Denoises skeleton latents (i.e., estimates added noise in the forward diffusion)  & \multirow{3}{*}{\makecell{Noise-sensitive performance}} \\
        & using \textbf{reverse diffusion}, conditioned on text embeddings, &\\
        & to naturally align both modalities in a \textbf{unified latent space} & \\
        \Xhline{2\arrayrulewidth}
    \end{tabular}}
    \vspace{-0.2cm}
    \caption{Comparison with ours TDSM with existing ZSAR methods.}
    \label{tab:novelty}
\end{table*}

\section{Detailed Structure of the Diffusion Transformer}

\subsection{About the Diffusion Transformer Design}

Note that our main contribution does not lie in the design of new components, but the first diffusion-based framework that is built upon DiT \cite{DiT} and MMDiT \cite{SD3} that have been well-validated for cross-modality alignment. Unlike the original DiT, we replace the class label embedding with $\mathbf{z}_g$ for semantic conditioning. Also shown in Table~\ref{tab:unet}, we also experimented with a U-Net backbone \cite{Unet}, but found DiT to perform better in our setting.

\begin{table}[tbp]
    \scriptsize
    \centering
    \resizebox{0.85\columnwidth}{!}{
    \def\arraystretch{1.2}
    \begin{tabular} {l|c|c|c|c}
        \Xhline{2\arrayrulewidth}
        \multirow{2}{*}{Backbone} & \multicolumn{2}{c|}{NTU-60\;(Acc, \%)} & \multicolumn{2}{c}{NTU-120 \;(Acc, \%)} \\
        \cline{2-5}
        & 55/5 split & 48/12 split & 110/10 split & 96/24 split \\
        \hline
        U-Net \cite{Unet} & 82.40 & 51.12 & 70.03 & 59.77 \\
        DiT (\textbf{TDSM}) & \textbf{\color{red}{86.49}} & \textbf{\color{red}{56.03}} & \textbf{\color{red}{74.15}} & \textbf{\color{red}{65.06}} \\
        \Xhline{2\arrayrulewidth}
    \end{tabular}}
    \vspace{-0.2cm}
    \caption{Comparison with ours TDSM with existing ZSAR methods.}
    \label{tab:unet}
\end{table}

\subsection{Diffusion Transformer Architecture}

The Diffusion Transformer $\mathcal{T}_{\text{diff}}$ takes $\mathbf{z}_{x,t}$, $\mathbf{z}_{g}$, $\mathbf{z}_{l}$, and $t$ as inputs (Fig.~\ref{fig:train} in the main paper). These inputs are embedded into corresponding feature representations $\mathbf{f}_{x,t}$, $\mathbf{f}_{c}$, and $\mathbf{f}_{l}$ as follows:
\begin{equation}
    \begin{split}
        &\mathbf{f}_{x,t} = \mathsf{Linear}(\mathbf{z}_{x,t}) + \mathsf{PE}_{x}, \\
        &\mathbf{f}_{c} = \mathsf{Linear}(\mathsf{TE}_{t}) + \mathsf{Linear}(\mathbf{z}_{g}),\\
        &\mathbf{f}_{l} = \mathsf{Linear}(\mathbf{z}_{l}) + \mathsf{PE}_{l},\\
    \end{split}
\end{equation}
where $\mathsf{PE}_{x}$ and $\mathsf{PE}_{l}$ are positional embeddings applied to the feature maps, capturing spatial positional information, while $\mathsf{TE}_{t}$ is a timestep embedding \cite{embedding} that maps the scalar $t$ to a higher-dimensional space. The embedded features $\mathbf{f}_{x,t}$, $\mathbf{f}_{c}$, and $\mathbf{f}_{l}$ are then passed through $B$ CrossDiT Blocks, followed by a Layer Normalization ($\mathsf{LN}$) and a final $\mathsf{Linear}$ layer to predict the noise $\hat{\bm{\epsilon}} \in \mathbb{R}^{M_{x} \times C}$.

\subsection{CrossDiT Block}

\begin{figure}[tbp]
  \centering
  \includegraphics[width=1.0\columnwidth]{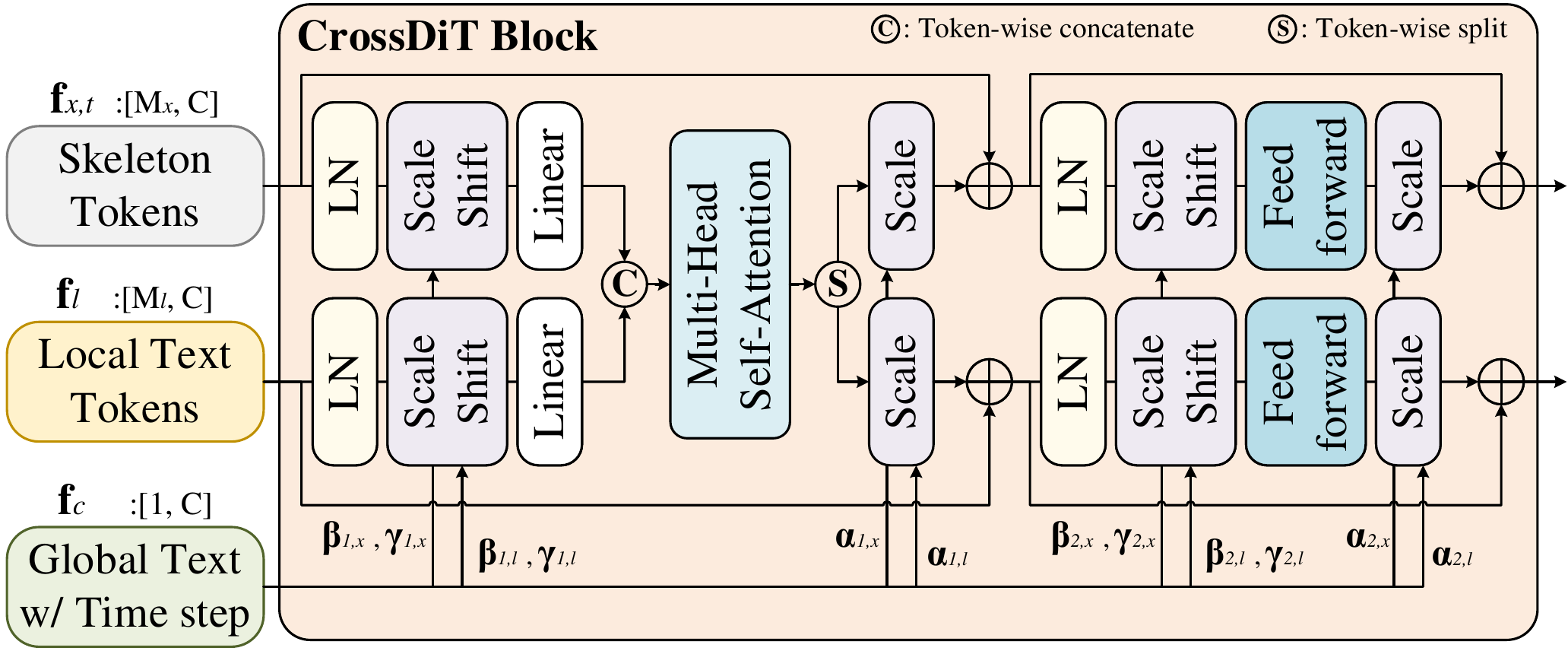}
  \caption{A detail structure of our CrossDiT Block.}
  \label{fig:block}
\end{figure}

The CrossDiT Block facilitates interaction between skeleton and text features, enhancing fusion through effective feature modulation \cite{SFTlayer, FiLM} and multi-head self-attention \cite{Attention}. Fig.~\ref{fig:block} shows a detail structure of our CrossDiT Block. Built upon the DiTs architecture \cite{DiT, SD3}, it leverages modulation techniques and self-attention mechanisms to efficiently capture the dependencies across these modalities. The skeleton feature $\mathbf{f}_{x}$ and local text feature $\mathbf{f}_{l}$ are first modulated separately using $\mathsf{Scale}$-$\mathsf{Shift}$ and $\mathsf{Scale}$ operations as:

\begin{equation}
    \begin{split}        &\left[\bm{\alpha}_{x}\mid\bm{\beta}_{x}\mid\bm{\gamma}_{x}\:|\:\bm{\alpha}_{l}\mid\bm{\beta}_{l}\mid\bm{\gamma}_{l}\right] = \mathsf{Linear}(\mathbf{f}_{c}),\\
    &\mathsf{Scale}\text{-}\mathsf{Shift}:\;\mathbf{f}_{i} \leftarrow (1 + \bm{\gamma}_{i}) \odot \mathbf{f}_{i} + \bm{\beta}_{i},\\
    &\mathsf{Scale}:\;\mathbf{f}_{i} \leftarrow \bm{\alpha}_{i} \odot \mathbf{f}_{i},\\
    \end{split}
\end{equation}

\noindent where $i \in \left\{x, l \right\}$ denotes the skeleton or local text feature, respectively. The parameters $\bm{\alpha}$, $\bm{\beta}$, and $\bm{\gamma}$ are conditioned on the global text feature $\mathbf{z}_{g}$ and timestep $t$, allowing the block to modulate feature representations effectively. Also, we compute query, key, and value matrices for both skeleton and local text features separately:

\begin{equation}
    [\mathbf{q}_{i}\mid\mathbf{k}_{i}\mid\mathbf{v}_{i}] = \mathsf{Linear}(\mathbf{f}_{i}).
\end{equation}

\noindent These matrices are token-wise concatenated and fed into a multi-head self-attention module, followed by a split to retain token-specific information as:

\begin{equation}
    [\mathbf{f}_{x}\mid\mathbf{f}_{l}] \leftarrow \mathsf{SoftMax}\left(\left[\mathbf{q}_{x}\mid\mathbf{q}_{l}\right]\left[\mathbf{k}_{x}\mid\mathbf{k}_{l}\right]^{\mathsf{T}}\right)\left[\mathbf{v}_{x}\mid\mathbf{v}_{l}\right].
\end{equation}

\noindent By leveraging the attention from skeleton, timestep, and text features, the CrossDiT Block ensures efficient interaction between modalities, promoting the skeleton-text fusion for discriminative feature learning and improved generalization to unseen actions.

\section{Implementation Details}
\label{sec:c}

Table~\ref{tab:notation} provides a detailed summary of the variables used in TDSM. We utilized $B=12$ CrossDiT Blocks, each containing a multi-head self-attention module with 12 heads. All feature dimensions were set to $C=768$. The local text features contained $M_{l}=35$ tokens, while the skeleton features were represented by a single token $M_{x}=1$. To ensure reproducibility, the random seed was fixed at 2,025 throughout all experiments. Skeleton features ($\mathbf{z}_x$) are extracted using skeleton encoder (Shift-GCN \cite{ShiftGCN} or ST-GCN \cite{STGCN}), resulting in a channel dimension of 256. For text features, two descriptions per action are encoded using the CLIP \cite{CLIP, openclip} text encoder, producing features with a channel dimension of 1,024. These features are concatenated along the channel dimension to form a unified text representation. 

\noindent \textbf{Fair comparison.} For the SynSE and PURLS settings, we utilize the same encoders as prior works to maintain consistency. We also encode $\mathbf{X}$ into $\mathbf{z}_x$ with $M_{x}=1$ to avoid any advantage from higher-resolution features (e.g., $M_{x}=T\times V$), again to ensure fair evaluation. These are a common practice in ZSAR task. For fair comparison, we used the same text prompts employed in existing works. When publicly available text prompts were provided, we used them as they were and did not heavily modify or augment them. For datasets without text descriptions (e.g., PKU-MMD \cite{PKUMMD}), we used GPT-4 \cite{GPT} to generate single description per action, ensuring consistency with the existing text styles.

\noindent \textbf{Hyper-parameter.} We tuned hyper-parameters extensively on the NTU-60 SynSE benchmark and then applied the same settings to all other datasets. Our method still achieved SOTA results.

\begin{table}[tbp]
\scriptsize
    \centering
    \setlength\tabcolsep{7pt}
    \resizebox{0.75\columnwidth}{!}{
    \def\arraystretch{1.25}
    \begin{tabular} {c|c|c}
        \Xhline{2\arrayrulewidth}
        & Module & Output Shape \\
        \hline
        $\mathbf{X}$ & & $T \times V \times M \times 3$ \\
        \hline
        $\mathbf{z}_x$ & $\mathcal{E}_x$ & $M_x \times 256$ \\
        \hline
        $\mathbf{f}_{x,t}$ & $\mathbf{z}_x$ Embed & $M_x \times 768$ \\
        \hline
        $\mathbf{z}_g$ & \multirow{2}{*}{$\mathcal{E}_d$} & $1 \times 1024$ \\
        $\mathbf{z}_l$ & & $M_l \times 1024$ \\
        \hline
        \multirow{2}{*}{$\mathbf{f}_{c}$} & \multirow{2}{*}{\makecell{$t$ Embed\\$\mathbf{z}_g$ Embed}} & \multirow{2}{*}{$1 \times 768$} \\
        & & \\
        \hline
        $\mathbf{f}_{l}$ & $\mathbf{z}_l$ Embed & $M_l \times 768$ \\
        \hline
        $\bm\epsilon$, $\hat{\bm\epsilon}$ & & $M_x \times 256$ \\
        \Xhline{2\arrayrulewidth}
    \end{tabular}}
    \caption{The details of feature shape.}
  \label{tab:notation}
\end{table}

\section{Additional Related Work}

\subsection{Skeleton-based Action Recognition}

Traditional skeleton-based action recognition assumes fully annotated training and test datasets, in contrast to other skeleton-based action recognition methods under zero-shot settings which aim to recognize unseen classes without explicit training samples. Early methods \cite{VALSTM, STLSTM, CoLSTM, IndRNN} employed RNN-based models to capture the temporal dynamics of skeleton sequences. Subsequent studies \cite{Ske2Grid, TACNN, MTCNN, PoseC3D} explored CNN-based approaches, transforming skeleton data into pseudo-images. Recent advancements leverage graph convolutional networks (GCNs) \cite{CTRGCN, FRHead, HDGCN, InfoGCN, LST, MSG3D, stcnet, BlockGCN} to effectively represent the graph structures of skeletons, comprising joints and bones. ST-GCN \cite{STGCN} introduced graph convolutions along the skeletal axis combined with 1D temporal convolutions to capture motion over time. Shift-GCN \cite{ShiftGCN} improved computational efficiency by implementing shift graph convolutions. Building on these methods, transformer-based models \cite{Hypergraph, IGFormer, skeletr, 3mformer, SkateFormer} have been proposed to address the limited receptive field of GCNs by capturing global skeletal-temporal dependencies. In this work, we adopt ST-GCN \cite{STGCN} and Shift-GCN \cite{ShiftGCN} to extract skeletal-temporal representations from skeleton data, transforming input skeleton sequences into a latent space for further processing in the proposed framework.

\clearpage